\documentclass[sigconf,natbib=true,anonymous=false,screen]{acmart}
\pdfoutput=1
\usepackage{amsmath}
\usepackage{amsthm}
\usepackage[ruled,linesnumbered,vlined]{algorithm2e}
\SetKwInOut{Input}{Input}
\SetKwInOut{Output}{Output}
\usepackage{algorithmic}
\usepackage{cleveref}
\usepackage{colortbl}
\usepackage{enumitem}
\setlist{leftmargin=*}

\usepackage{xcolor}
\usepackage{graphicx}

\usepackage{caption}
\usepackage{subcaption}  % enable subfigure
\usepackage{mwe}

\usepackage{multirow}

\usepackage{mdframed}

\usepackage{bbding}

\usepackage{threeparttable}

\usepackage[ruled,linesnumbered,vlined]{algorithm2e}
\SetKwInOut{Input}{Input}
\SetKwInOut{Output}{Output}
\usepackage{algorithmic}
\usepackage{graphicx}
\begin{document}

\title{Out-of-distribution Rumor Detection via Test-Time Adaptation}

\author[Xiang Tao]{Xiang Tao$^{1,2,*}$, Mingqing Zhang$^{1,2,*}$, Qiang Liu$^{1,2}$, Shu Wu$^{1,2}$, Liang Wang$^{1,2}$}
\affiliation{
	\institution{$^1$Center for Research on Intelligent Perception and Computing, Institute of Automation, Chinese Academy of Sciences}
	\institution{$^2$School of Artificial Intelligence, University of Chinese Academy of Sciences}
	\country{}
}
\email{{xiang.tao,mingqing.zhang}@cripac.ia.ac.cn, {qiang.liu,shu.wu,wangliang}@nlpr.ia.ac.cn}

\thanks{*The first two authors made equal contribution to this work.}
%\dag To whom correspondence should be addressed.}
\renewcommand{\shortauthors}{Xiang et al.}

\newcommand{\themodel}{\textsf{AUG-MAE}\xspace}
\newcommand{\themodelm}{\textsf{$\text{\themodel}_{\text{MoE}}$}\xspace}

\begin{abstract}
Due to the rapid spread of rumors on social media, rumor detection has become an extremely important challenge. Existing methods for rumor detection have achieved good performance, as they have collected enough corpus from the same data distribution for model training. 
However, significant distribution shifts between the training data and real-world test data occur due to differences in news topics, social media platforms, languages and the variance in propagation scale caused by news popularity. This leads to a substantial decline in the performance of these existing methods in Out-Of-Distribution (OOD) situations. To address this problem, we propose a simple and efficient method named \underline{T}est-time \underline{A}daptation for \underline{R}umor \underline{D}etection under distribution shifts (TARD). This method models the propagation of news in the form of a propagation graph, and builds propagation graph test-time adaptation framework, enhancing the model’s adaptability and robustness when facing OOD problems. Extensive experiments conducted on two group datasets collected from real-world social platforms demonstrate that our framework outperforms the state-of-the-art methods in performance.
% is also incapable of ensuring the uniformity of the learned representations.
\end{abstract}

%%
%% The code below is generated by the tool at http://dl.acm.org/ccs.cfm.
%%


% \received{20 February 2007}
% \received[revised]{12 March 2009}
% \received[accepted]{5 June 2009}

\maketitle

\section{Introduction}

The growth of social media has accelerated the diffusion of news information. However, this also poses certain risks, such as the spread of rumors that can affecting people's lives and the stability of society~\cite{harm-rumors,fmisinformaton,health-sear,spreader}. 
Traditional rumor detection methods primarily rely on extensive corpora collected from specific domains, ensuring the training of models that perform well under controlled conditions~\cite{content-based1,pro-stru1,pro-stru2,RNN4R,RNN4R-2,CNN-based,BIGCN,GACL,sig21}.

\begin{figure*}[htb]
    \centering
    \includegraphics[width=\linewidth]{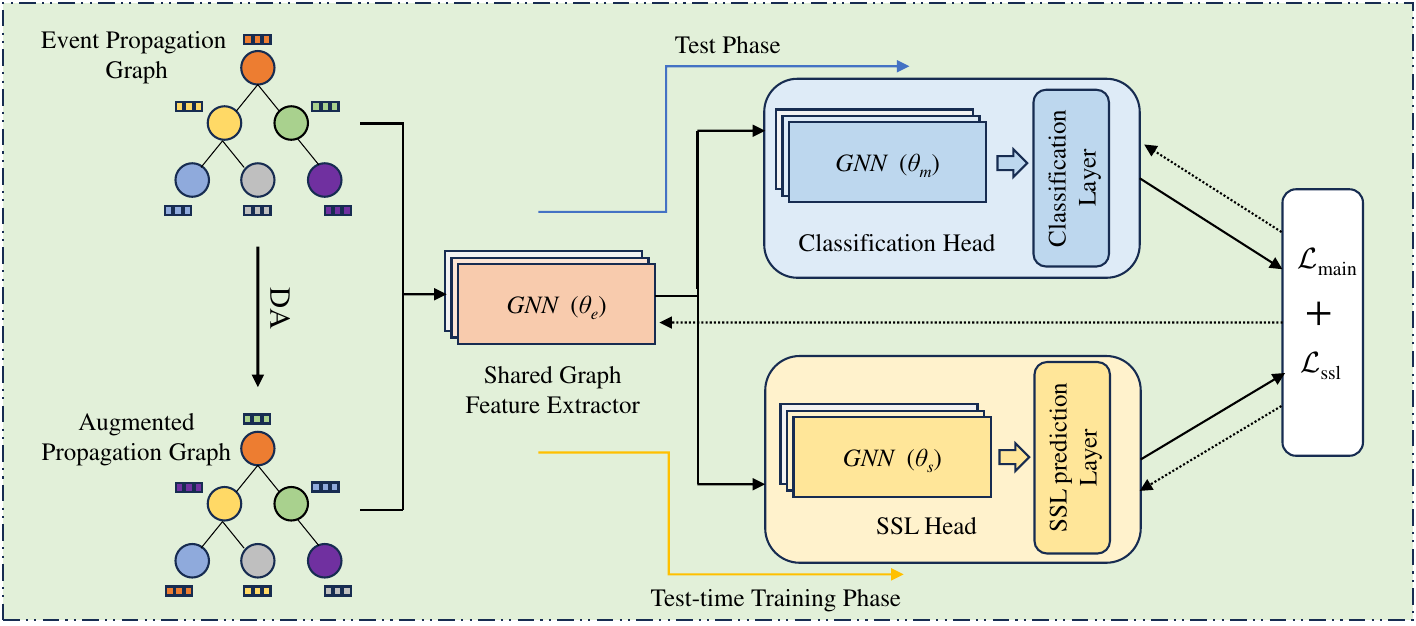}
    \caption{The overall framework of our proposed TARD model.During the training phase, all model parameters are updated. In the test-time training phase, only the parameters of the shared feature extractor and the SSL head are updated, allowing the model to adapt to the data distribution of the test dataset. During the testing phase, the updated shared feature extractor and classification head are used to classify the test dataset.}
    \label{overview}
\end{figure*}

However, these methods often struggle in Out-Of-Distribution (OOD) scenarios, primarily due to the dynamic and diverse nature of social media content. For example, different news topics, such as politics, technology, and entertainment, may display significant variations in language style, user interaction patterns, and propagation speeds. Additionally, various social media platforms like Twitter, Facebook, and Weibo differ in user demographics, communication styles, and content presentation formats. The propagation scale and speed of both popular and less popular news stories can vary significantly, posing a challenge to the model's generalizability. Furthermore, users from diverse linguistic and cultural backgrounds may exhibit differences in expression styles and vocabulary habits~\cite{zero-shot, naacl}. 
This diversity can affect the presentation of rumors and the way users respond to it.
These factors contribute to a significant distribution shift between the training data and the test data encountered in real-world scenarios. This discrepancy leads to a marked decrease in the effectiveness of existing rumor detection models when dealing with OOD data.

Inspired by the latest advancements in OOD work in computer vision and graph neural networks~\cite{TTTcv,TTTGNN,ttt++}, we propose a simple and efficient method named \underline{T}est-time \underline{A}daptation for \underline{R}umor \underline{D}etection under distribution shifts (TARD) to address this challenge. 
The primary goal of this framework is to significantly enhance the model's adaptability and robustness, especially when dealing with OOD problems frequently encountered in the diverse and evolving social media environment.
Specifically, we developed a test-time adaptation framework for the news propagation graph. This framework introduces an additional self-supervised learning channel to the base of supervised rumor classification learning, employing a three-step training process: training, test-time training, and testing. The model's architecture is a Y-structure with a shared bottom and two branches (classification head and self-supervised head). During the training phase, our model undergoes joint optimization for both supervised and unsupervised tasks, enabling it to capture a comprehensive understanding of the underlying data. In the subsequent test-time training phase, we freeze the main task head and only leverage the shared feature extractor and self-supervised learning task to extract latent features from the test samples. These features act as crucial cues, guiding the recalibration of our model to better adapt to the nuanced data distribution observed in the test samples. In the test phase, we only use the shared layer and classification layer to detection rumors. 

Furthermore, a direct application of test-time training presents challenges. It may inadvertently lead to model overfitting to the self-supervised learning task of specific test samples, resulting in substantial representational distortion. This distortion, in turn, could hinder the model's performance on the primary rumor classification task. To address this potential issue, we introduce an adaptive constraint to the objective function during the test-time training phase. The core idea is to constrain the embedding space output of the shared feature extractor between training and test samples, preventing significant fluctuations in representation.

Extensive experiments conducted on two real-world datasets confirm that our TARD method outperforms state-of-the-art approaches in overall performance.
The main contributions of our work are outlined as follows:
\begin{itemize}
    \item We propose a TARD framework to address the challenges faced by rumor detection models in OOD situations.
    \item We conduct extensive experiments on several real-world datasets in the OOD experimental setup. The results indicate that our rumor detection framework demonstrates excellent performance in OOD situations.
\end{itemize}

\section{The proposed TARD Model}

\subsection{Problem formulation}
The problem is defined as a classification task, where the objective is to learn a classifier that can accurately detect rumors.
Specifically, for a given rumor dataset $\mathcal{C} =\{C_1, C_2, ..., C_M\}$, where $C_i$ is the $i$-th event.
We defined each event $C_i = \{r, w_1,..., w_{{N_i}-1}, \boldsymbol{\mathcal{G}}_i\}$, where $N_i$ is the number of posts, $r$ refers to the source post, each $w_j$ represents the $j$-th relevant responsive post, and $\boldsymbol{\mathcal{G}}_i$ defined as a graph represents the propagation structure of $C_i$. 
The graph $\boldsymbol{\mathcal{G}}_i = \{\boldsymbol{V}_i,\boldsymbol{A}_i,\boldsymbol{X}_i\}$, where $\boldsymbol{V}_i$ refers to the set of nodes corresponding to $N_i$ posts and $\boldsymbol{A}_i\in\{0,1\}^{N_i\times N_i}$ as an adjacency matrix where:
\begin{equation}  
    a_{st}^i = \left\{\begin{array}{cc}
         1, & \text{if } e_{st}^i\in{E_i}  \\
         0, & \text{otherwise},
    \end{array}\right.
\end{equation}
where $E_i = \{e_{st}^i|s,t \in \{0,1,...,N_i-1\}\}$ represents the set of edges connecting a post to its responsive posts.
$\boldsymbol{X}_i = [ x_0,x_1,...,x_{N_i-1} ]^T$ denote a node feature matrix extracted from the posts in $C_i$.
We adopt the same approach as ~\cite{naacl} by using the XLM-RoBERTa~\cite{roberta} to semantically align and encode the source and comments to form the feature matrix $\boldsymbol{X}_i$. Besides, each event $C_i$ is labeled with a ground-truth label $y_i$. 
% Here, we define the problem statement as follows:

% \textbf{Rumor Detection}: The task is to develop a classifier, denoted as $f: C_i \xrightarrow{} y_i$, where $C_i$ represents a event of rumor dataset with their corresponding graph structure and textual features. 

\subsection{Framework}
As \cref{overview} shows, our TARD model architecture is a Y-structure with a shared bottom and two branches. The shared bottom refers to a shared feature extractor, consisting of $l$-layer of GNN with its parameters denoted as $\boldsymbol{\Theta_e}$. One of the branches is the classification head, which is used for the main task of rumor classification. It consists of $p$-layer of GNN with its parameters denoted as $\boldsymbol{\Theta_m}$. 
While the other branch is the self-supervised learning head. It consists of $t$-layer of GNN with its parameters denoted as $\boldsymbol{\Theta_s}$. 

\subsubsection{Self-supervised Learning Task}
\label{ssl-text}
In our TARD framework, the design of self-supervised learning task is crucial because we aim to effectively capture the data distribution through it, enabling the model to correct itself during the test-time training phase. Inspired by \cite{TTTGNN}, we employ graph contrastive learning as our self-supervised learning task.
Specifically, we perform a random shuffling of posts in the propagation graph of the original view $\boldsymbol{\mathcal{G}_0}$ to yields our augmented view $\boldsymbol{\mathcal{G}_1}$, where node features are randomly shuffled across all nodes within the graph. The original view and the augmented view are then input into the shared feature extractor and the SSL head to obtain their representations $\boldsymbol{H_0}$ and $\boldsymbol{H_1}$. Through pooling operations, a global graph representation $\boldsymbol{g_0}$ can be summarized from the node representation $\boldsymbol{H_0}$ of the original view $\boldsymbol{\mathcal{G}_0}$. 
Then we treat the representations of each node representation in the original graph, and the graph representation, as positive samples. The representations of each node in the augmented graph, and the graph representation, as negative samples. 
We set $\mathcal{D}(\boldsymbol{H_{ji},g_0}) = Sigmoid(\boldsymbol{H_{ji} * g_0})$, where $\boldsymbol{H_{ji}}$ denotes the representation of node $i$ from view $\boldsymbol{\mathcal{G}_j}$ and $*$
denotes inner product. 

The parameters of shared feature extractor and SSL head can be learned by
\begin{equation}
     \mathcal{L}_{\text{s}} = -\frac{1}{2N}(\sum_{i=1}^{N}(log\mathcal{D}(\boldsymbol{H_{0i},g_0}) + log(1 - \mathcal{D}(\boldsymbol{H_{1i},g_0})))
\end{equation}
where $N$ denotes the number of nodes in the input graph.

\subsubsection{Train Phase}
The training phase can update the model parameters based on both the main task and self-supervised task. Specifically, given the training data $\boldsymbol{\mathcal{G}_0}$, we input both the original view and the augmented view into a shared feature extractor. Subsequently, we perform learning for both the main task and the self-supervised task. The main task involves supervised classification of the original graph. 
The supervised loss of main task can be expressed as:
\begin{equation}
    \mathcal{L}_{\text{m}} = -\frac{1}{N} \sum_{k=1}^{N} \sum_{j=1}^{C}\boldsymbol{y}_{k,j}log(\boldsymbol{\hat{y}}_{k,j}),
    \label{sup-loss}
\end{equation}
where $\boldsymbol{y}_{k,j}$ denotes ground-truth label that has been one-hot encoded. and $\boldsymbol{\hat{y}}_{k,j}$ denotes the predicted probability distribution of event index $k \in \{1,2...,N\}$ belongs to class $j \in \{1,2,...C\}$.

For the self-supervised task, the representations of the original view and augmented view are fed into the SSL head for self-supervised learning as described in \ref{ssl-text}. Our model can be learned by:
\begin{equation}
\begin{aligned}
    \boldsymbol{\Theta_e^{\star}, \Theta_m^{\star}, \Theta_s^{\star}} = &\arg \min_{\Theta_e, \Theta_m,\Theta_s}(\mathcal{L}_m(\boldsymbol{\mathcal{G}_0; \Theta_e, \Theta_m}) \\ &+\alpha_1\mathcal{L}_s(\boldsymbol{\mathcal{G}_0,\mathcal{G}_1; \Theta_e, \Theta_s}))
    \label{thetafinal} 
\end{aligned}
\end{equation}
where $\alpha_1$ is adjustable hyperparameter used to control the weight of the self-supervised loss.

\subsubsection{Test-Time Training Phase}
During the Test-time training phase, our objective is to recalibrate the model parameters through a self-supervised task to adapt to the data distribution of the test set. Specifically, after the completion of training, the main task classification head is frozen. When inputting the test data $\boldsymbol{\mathcal{G}_0}$, we obtain its augmented view $\boldsymbol{\mathcal{G}_1}$ and feed both views into the shared feature extractor and the SSL head for self-supervised learning. It is important to note that at this stage, only the parameters of the shared feature extraction layer and the SSL task head are updated by:

\begin{equation}
    \boldsymbol{\Theta_e^{\star},  \Theta_s^{\star}} = \arg \min_{\Theta_e, \Theta_s}\mathcal{L}_s(\boldsymbol{\mathcal{G}_0,\mathcal{G}_1; \Theta_e, \Theta_s})
    \label{thetafinal}
\end{equation}

\subsubsection{Adaptation Constraint}
During the test-time training phase, when recalibrating the model through self-supervised learning tasks, without imposing constraints, it may lead to overfitting on the self-supervised tasks, resulting in a significant distortion of the learned representations. Inspired by \cite{TTTGNN}, we introduce an adaptive feature constraint term during the model's test-time training, to prevent this phenomenon. This is done to avoid representation distortion. Therefore, the objective defined by \cref{thetafinal} is finall updated as:

\begin{equation}
    \boldsymbol{\Theta_e^{\star},  \Theta_s^{\star}} = \arg \min_{\Theta_e, \Theta_s}(\mathcal{L}_s(\boldsymbol{\mathcal{G}_0,\mathcal{G}_1; \Theta_e, \Theta_s}) + \alpha_2\mathcal{L}_c(\boldsymbol{\mu,\eta,\mu_t,\eta_t}))
    \label{thetafinal}
\end{equation}
where $\alpha_2$ is adjustable hyperparameters used to control the weight of the constraint term. In $\mathcal{L}_c$, $\boldsymbol{\mu}$ and $\boldsymbol{\eta}$ are the vector mean and covariance matrix of the node feature matrix obtained after the training data passes through the shared feature extractor upon completion of the training stage. Whereas $\boldsymbol{\mu_t}$ and $\boldsymbol{\eta_t}$ are the vector mean and covariance matrix obtained after the test data passes through the shared feature extractor during the test-time training stage.

\subsubsection{Test Phase}

When a test data undergoes test-time training, we immediately test it, then input the next test data for test-time training and testing. During testing, the recalibrated shared feature extractor is applied to the test dataset, and the output is fed into the main task classification head for testing.

\section{Experiments}
In this section, we first evaluate the effectiveness of our proposed TARD model  by comparing it with several baseline models, and give some discussion and analysis. 
Next, we conducted ablation study to analyze and evaluate the effectiveness of each module in TARD.
Finally, we perform a sensitivity analysis of the hyper-parameters in TARD, discussing the impact of each hyperparameter on the experimental results.
\subsection{Experimental Setups}
\subsubsection{\textbf{Datasets}}

We evaluated the TARD model using two group of real-world OOD rumor datasets. Both datasets exhibit distribution shifts in terms of event topics, social media platforms, and language between the training and testing sets. The first group is Twitter dataset~\cite{twitter} and Weibo-Covid19 dataset~\cite{naacl}. The second group is the Weibo dataset~\cite{RNN4R-2} and the Twitter-Covid19 dataset~\cite{naacl}. Twitter and Twitter-COVID19 are English rumor datasets with conversation thread in tweets while Weibo and Weibo-COVID19 are Chinese rumor datasets with the similar composition structure. These datasets have two tags: Rumor and Non-Rumor, used for binary classification of rumors and non-rumors. 

\subsubsection{\textbf{Baselines}}

We compare TARD with state-of-the-art
 rumor detection models, including: 
(1) $RvNN$~\cite{ma2018rumor} is a model based on tree-structured recursive neural networks.
(2) $PLNA$~\cite{khoo2020interpretable} is a transformer-based model for rumor detection.
(3) $BiGCN$~\cite{bian2020rumor} is a GCN-based model based on directed conversation trees.
(4) $GACL$~\cite{GACL} is a GNN-based model using adversarial and contrastive learning.

\begin{table*}
    \centering
    \caption{Rumor detection results on the target test datasets.}
    \vspace{-5pt}
    \label{results}
    \begin{threeparttable}
        \begin{minipage}{\linewidth}
            \centering
            \begin{tabular}{c|cc|cc|cc|cc}
                \toprule
                \ Target(Source) & \multicolumn{4}{c|}{Weibo-COVID19 (Twitter)}  & \multicolumn{4}{c}{Twitter-COVID19 (Weibo)} \\
                \midrule
                \multirow{2}*{Model} & \multirow{2}*{Acc.} & \multirow{2}*{Mac-F1} & Rumor & Non-Rumor & \multirow{2}*{Acc.} & \multirow{2}*{Mac-F1} & Rumor & Non-Rumor \\
                & & & F1 & F2 & & & F1 & F2 \\ 
                \midrule
                RvNN & 0.514 & 0.482 & 0.538 & 0.426 & 0.540 & 0.391 & 0.534 & 0.247 \\
                PLAN & 0.532 & 0.496 & 0.578 & 0.414 & 0.573 & 0.423 & 0.549 & 0.298 \\
                BiGCN & 0.569 & 0.508 & 0.586 & 0.429 & 0.616 & 0.415 & 0.577 & 0.252 \\
                GACL & 0.682 & 0.615 & 0.702 & 0.528 & 0.686 & 0.520 & 0.686 & 0.354 \\
                \rowcolor{gray!25}
                \midrule
                TARD & \textbf{0.737} & \textbf{0.727} & \textbf{0.780} & \textbf{0.673} & \textbf{0.715} & \textbf{0.652} & \textbf{0.800} & \textbf{0.504}  \\
                \bottomrule
            \end{tabular}
        \end{minipage}
    \end{threeparttable}
    \vspace{-10pt}
\end{table*}

\subsubsection{\textbf{Experimental Settings}}
In this work, we conduct rumor detection in OOD situations. Specifically, we use Weibo~\cite{RNN4R-2} (or Twitter~\cite{twitter}) datasets as the source data, and Twitter-COVID19~\cite{naacl} (or Weibo-COVID19~\cite{naacl}) datasets as the target. We use accuracy and macro-averaged $F1$, as well as class-specific $F1$ scores as the evaluation metrics. 
Furthermore, We adopt 1-layer GCN as the shared feature extractor, 1-layer GCN as the classification and 1-layer as the self-supervised head.
We set $\alpha_1 = 1, \alpha_2 = 0.1$ for Twitter, and $\alpha_1 = 0.01, \alpha_2 = 1$ for Weibo.

\subsection{Overall Performance}
\cref{results} shows the performance of our proposed method versus all the compared methods on the Weibo-COVID19 and Twitter-COVID19 test sets with pre-determined training datasets, where the bold part represents the best performance. The experimental results demonstrate that the proposed TARD performs exceptionally well among all baseline models, confirming the advantages of test-time adaptation framework integrated with self-supervised learning to learn the OOD information.\\
In the OOD experimental setup, due to significant distribution shift between the test set and training set, the performance of other models is significantly worse. Our proposed TARD method, after learning features from the training set, leverages a test-time training module to effectively enhance the model's generalization ability to the test set. Additionally, there is an additional constraint ensuring that the embeddings of test graph samples closely align with the distribution of embeddings generated for training graph samples under the guidance of the main task. This results in our proposed model achieving state-of-the-art performance.

\subsection{Ablation Study}
\begin{figure}
    \centering
    \includegraphics[width=0.7\linewidth]{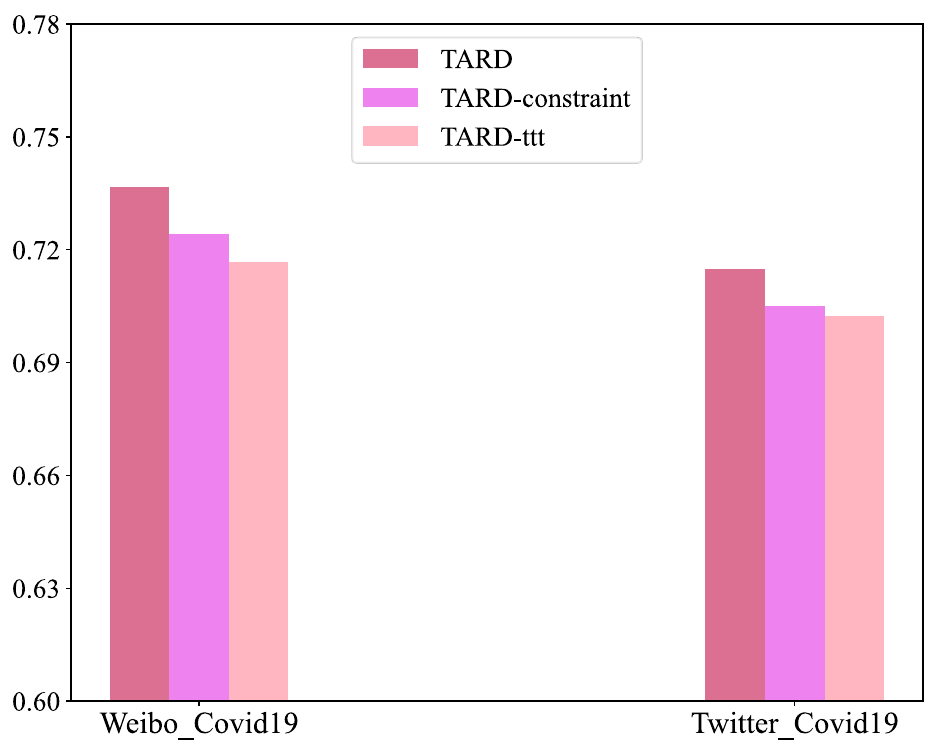}
    \vspace{-10pt}
       \caption{Ablation studies on our proposed model}
       \label{fig:ablation}
    \vspace{-15pt}
\end{figure}

To evaluate the efficacy of the various modules of TARD, we conduct a comparative analysis by comparing it with the following variants: (1) \textit{TARD-constraint} removes the adaptation constraint. (2) \textit{TARD-ttt} eliminates the test-time training modules, directly applying the features learned from source dataset to rumor detection in target dataset.
The results are summarized in \cref{fig:ablation}. We have the following observations from this figure:

\begin{enumerate}
    \item[1)] By comparing TARD and TARD-constraint, we can observe that the accuracy is reduced. Obviously, the introduction of the adaptation constraint, as opposed to directly employing test-time training on the test sample, resulted in a significant performance improvement.
    \item[2)] Removing the test-time training module results in a decrease in the model's performance, and its performance is lower than TARD-constraint, which includes test-time training module. This indicates that the test-time training module can effectively learn information from OOD samples, enhancing the model's performance.
\end{enumerate}

\begin{figure}
    \centering
    \includegraphics[width=0.98\linewidth]{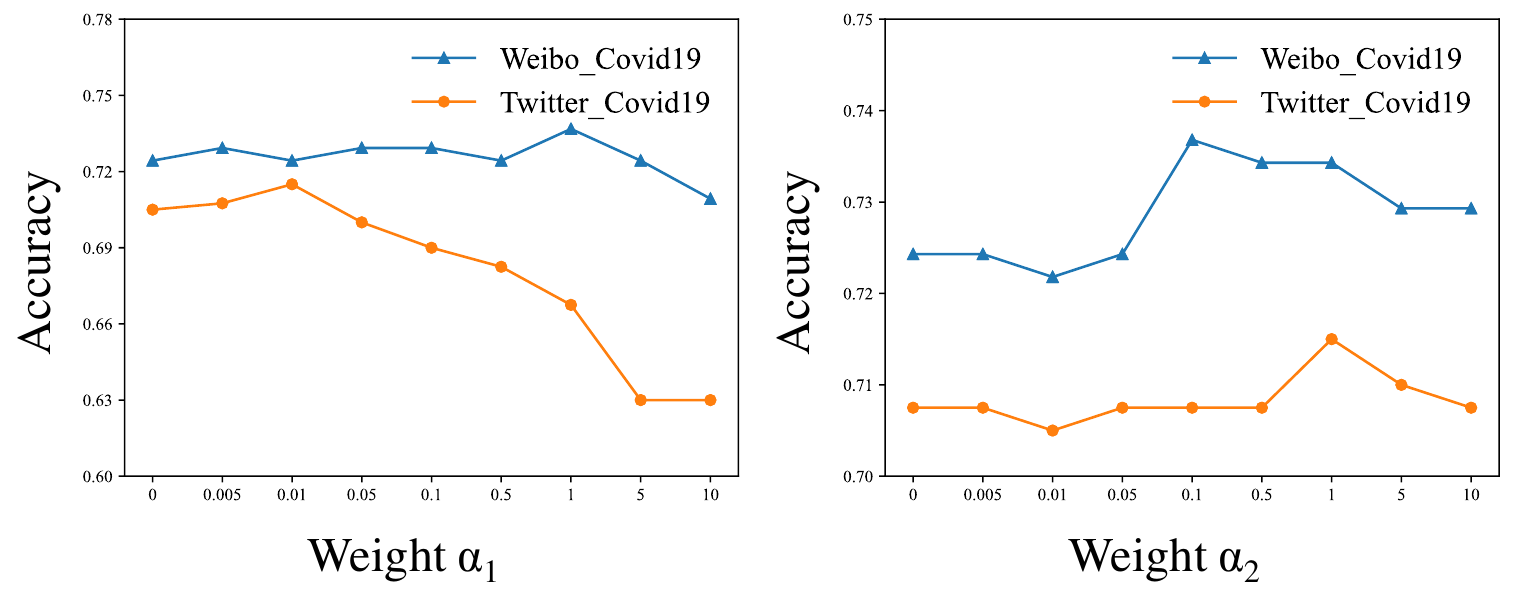}
    \vspace{-10pt}
       \caption{Sensitivity analysis of hyperparameters $\alpha_1$ and $\alpha_2$.}
       \label{fig:weight}
    \vspace{-10pt}
\end{figure}
\subsection{Sensitivity Analysis}

We conduct sensitivity analysis with hyper-parameters on the key designs of TARD by selecting nine data points between $0$ and $10$. \cref{fig:weight} shows the effect of varied hyper-parameter values, from which we have the following observations.

\subsubsection{\textbf{Effect of weight of self-supervised loss $\alpha_1$.}}

As shown in the left picture of ~\cref{fig:weight}, we observed that as the parameter $\alpha_1$ increases, the model's performance shows fluctuations and gradually improves. This is attributed to the inclusion of appropriate self-supervised learning, enhancing the model's generalization and robustness. However, when $\alpha_1$ exceeds a certain threshold, the model's performance starts to decline. This is because of overfitting of the model to the self-supervised learning task during training.

\subsubsection{\textbf{Effect of weight of adaptation constraint loss $\alpha_2$.}}

As shown in the right picture of ~\cref{fig:weight}, we observed that as $\alpha_2$ increases, the model's performance begins to fluctuate, and it achieves significant performance improvements when appropriate weight values are attained. When the weight becomes too large, however, the model's performance variation becomes minimal. This is because the model's learning is heavily constrained at this point, thereby impacting its performance.

\begin{comment}
\begin{table}[t]
\centering
\caption{ Statistics of the datasets}
\vspace{-10pt}
\label{datasets}
\begin{threeparttable}
\begin{tabular}{c|cc|cc}
\toprule
    \multirow{2}*{Statistics} & source& target&source&target\\
    & Twitter  & Weibo-COVID19 & Weibo &Twitter-COVID19  \\
    \midrule
    \# of claims & 1154 & 399 & 4649 & 400  \\
    \# of tree nodes & 60409 & 26687 & 1956449 & 406185  \\
    \# of non-rumors & 579 & 146 & 2336 & 148  \\
    \# of rumors & 575 & 253 & 2313 & 252  \\
    Avg.\# of posts/tree & 52 & 67 & 420 & 1015  \\
    
\bottomrule
\end{tabular}
\end{threeparttable}
\vspace{-10pt}
\end{table}
\end{comment}

\section{Conclusion}
In this paper, we propose a simple and effective test-time adaptation framework, named TARD. It addresses the significant performance degradation of existing rumor detection methods when facing OOD situations by constructing a news propagation graph and combining supervised learning with self-supervised learning.  This integration enhances the model's adaptability and robustness when dealing with OOD challenges. Extensive experiments conducted on two datasets collected from real-world social media platforms demonstrate that our approach surpasses the state-of-the-art methods in terms of performance in the OOD experimental setup.

% \begin{acks}
% To Robert, for the bagels and explaining CMYK and color spaces.
% \end{acks}

\clearpage
\balance
\bibliographystyle{ACM-Reference-Format}
\bibliography{paper}

\end{document}